\documentclass{article}

\usepackage[preprint]{corl_2026} %
\usepackage{amsmath}
\usepackage{amssymb}
\definecolor{gradA}{RGB}{150,185,130}
\definecolor{gradAB}{RGB}{180,175,125}
\definecolor{gradB}{RGB}{200,165,115}
\definecolor{gradBC}{RGB}{175,170,155}
\definecolor{gradC}{RGB}{135,165,200}
\definecolor{linkred}{rgb}{0.918,0.200,0.137}
\usepackage{cleveref}
\crefname{section}{Sec.}{Secs.}
\Crefname{section}{Sec.}{Secs.}
\crefname{subsection}{Sec.}{Secs.}
\Crefname{subsection}{Sec.}{Secs.}
\crefname{figure}{Fig.}{Figs.}
\Crefname{figure}{Fig.}{Figs.}
\usepackage{graphicx}
\usepackage{subcaption}
\usepackage{booktabs}
\usepackage[table]{xcolor}
\definecolor{teaiblue}{HTML}{1A50D1}
\hypersetup{citecolor=teaiblue,linkcolor=linkred,urlcolor=teaiblue}
\usepackage{todonotes}
\usepackage{tikz}
\usepackage[most]{tcolorbox}

\makeatletter
\renewcommand{\@maketitle}{%
  \vbox{%
    \hsize\textwidth
    \linewidth\hsize
    \vskip 0em
    \noindent\includegraphics[height=1cm]{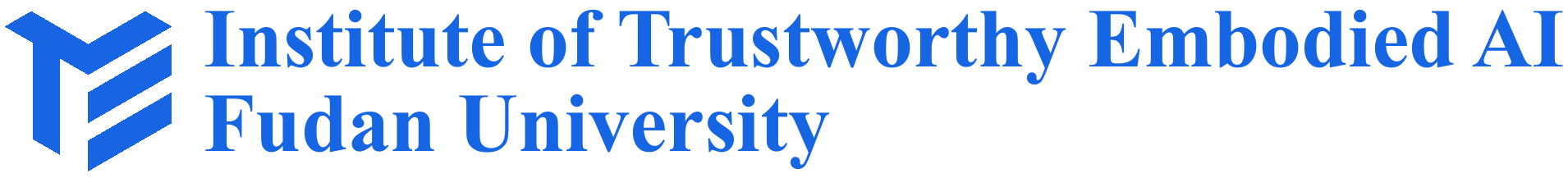}\par
    \vskip 0.3em
    {\color{teaiblue}\hrule height 3\p@}
    \vskip 1.8em
    \vskip -\parskip
    \centering
    {\LARGE\bf \@title\par}
    \vskip 2.1em
    \vskip -\parskip
    {\color{teaiblue}\hrule height 1.5\p@}
    \vskip 0.2em
    \if@conferencefinal
      \def\And{%
        \end{tabular}\hfil\linebreak[0]\hfil%
        \begin{tabular}[t]{c}\bf\rule{\z@}{12\p@}\ignorespaces%
      }
      \def\AND{%
        \end{tabular}\hfil\linebreak[4]\hfil%
        \begin{tabular}[t]{c}\bf\rule{\z@}{12\p@}\ignorespaces%
      }
      \begin{tabular}[t]{c}\bf\rule{\z@}{12\p@}\@author\end{tabular}%
    \else
      \if@preprinttype
        \def\And{%
          \end{tabular}\hfil\linebreak[0]\hfil%
          \begin{tabular}[t]{c}\bf\rule{\z@}{12\p@}\ignorespaces%
        }
        \def\AND{%
          \end{tabular}\hfil\linebreak[4]\hfil%
          \begin{tabular}[t]{c}\bf\rule{\z@}{12\p@}\ignorespaces%
        }
        \begin{tabular}[t]{c}\bf\rule{\z@}{12\p@}\@author\end{tabular}%
      \else
        \begin{tabular}[t]{c}\bf\rule{\z@}{12\p@}
          Anonymous Author(s) \\
          Affiliation \\
          Address \\
          \texttt{email} \\
        \end{tabular}%
      \fi
    \fi
    \vskip 2.2em \@minus 0.7em
  }
}
\renewcommand{\section}{%
  \@startsection{section}{1}{\z@}%
                {-2.0ex \@plus -0.5ex \@minus -0.2ex}%
                { 1.5ex \@plus  0.3ex \@minus  0.2ex}%
                {\large\bf\raggedright\color{teaiblue}}%
}
\renewcommand{\subsection}{%
  \@startsection{subsection}{2}{\z@}%
                {-1.8ex \@plus -0.5ex \@minus -0.2ex}%
                { 0.8ex \@plus  0.2ex}%
                {\normalsize\bf\raggedright\color{teaiblue}}%
}
\makeatother

\newcommand{\name}{ActiveMimic}

\title{\name: Egocentric Video Pretraining \\with Active Perception}

\def\onedot{.}
\def\eg{\emph{e.g}\onedot} %

\author{
  \textbf{Xingyao Lin}\textsuperscript{\rm 1,2} \quad
  \textbf{Guojin Zhong}\textsuperscript{\rm 1} \quad
  \textbf{Tianyi Lu}\textsuperscript{\rm 1} \quad \\
  \textbf{Ziyi Ye}\textsuperscript{\rm 1} \quad
  \textbf{Yichen Zhu}\textsuperscript{\rm 3} \quad
  \textbf{Zuxuan Wu}\textsuperscript{\rm 1,2,4} \quad
  \textbf{Yu-Gang Jiang}\textsuperscript{\rm 1} \quad \\
  \textsuperscript{\rm 1} Fudan University,
  \textsuperscript{\rm 2} Shanghai Innovation Institute,
  \textsuperscript{\rm 3} Current Robotics,
  \textsuperscript{\rm 4} NeoteAI
}

\begin{document}

\maketitle
\thispagestyle{plain}

\vspace{-3.0em}
\begin{figure*}[h]
    \centering
    \includegraphics[width=\linewidth]{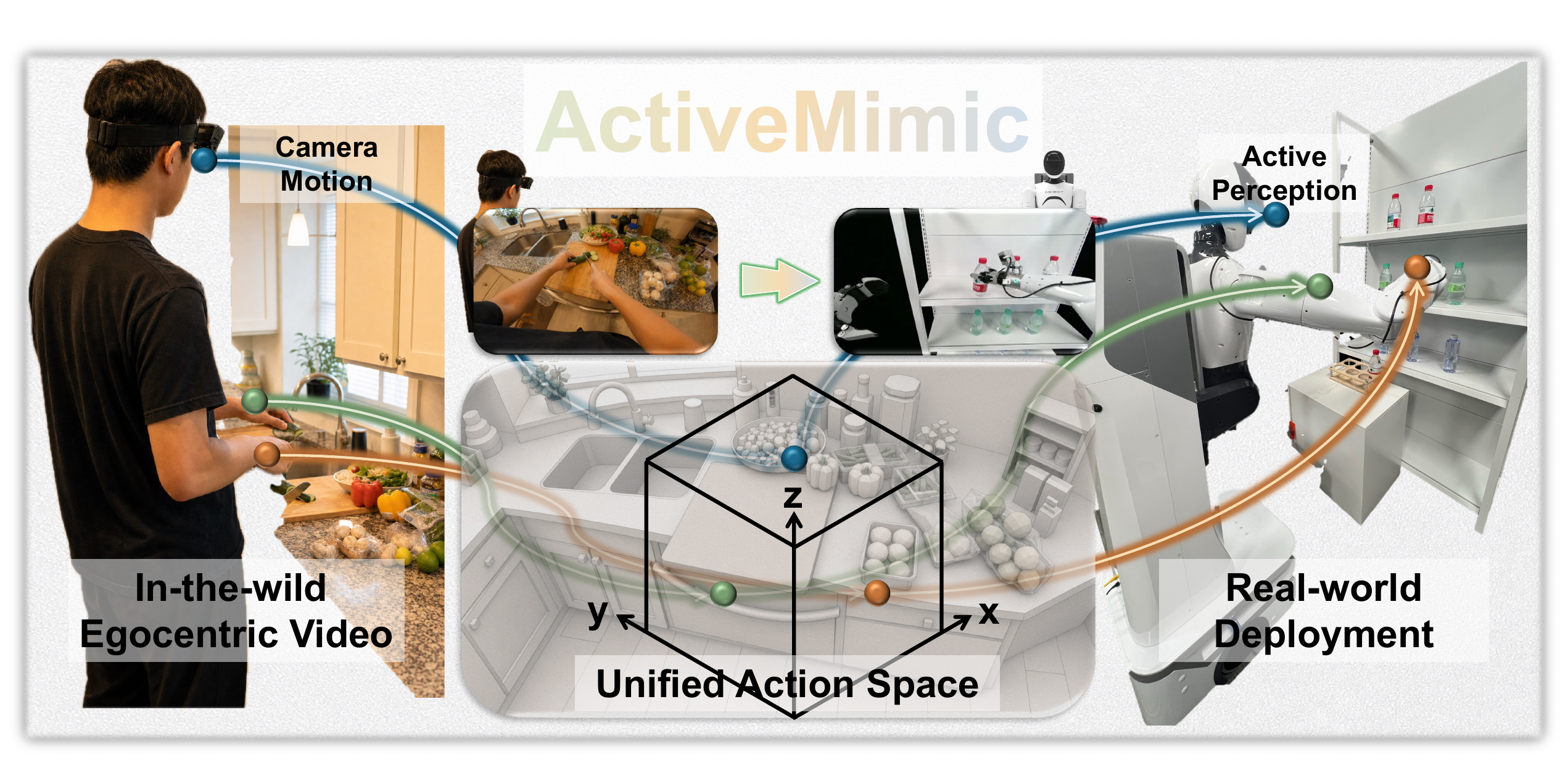}
    \caption{\textbf{\name{} acquires active perception from in-the-wild egocentric human video and transfers it to real-world humanoid robots.} \textit{\textcolor{gradA}{Left} \textcolor{gradAB}{to} \textcolor{gradB}{center:}} egocentric camera motion and wrist action together form a 27-dimensional unified action representation that enables the model to jointly learn active perception and manipulation. \textit{\textcolor{gradB}{Center} \textcolor{gradBC}{to} \textcolor{gradC}{right:}} active perception is transferred to a humanoid robot, which repositions its viewpoint actively during task execution.}
    \label{fig:teaser}
\end{figure*}

\vspace{-1.0em}
\begin{tcolorbox}[
  colback=teaiblue!7,
  colframe=white,
  boxrule=0pt,
  arc=2.5mm,
  left=4mm,
  right=4mm,
  top=4mm,
  bottom=4mm
]
  \setlength{\parindent}{0pt}
  \setlength{\parskip}{0pt}
  \begin{center}
    {\normalsize\bfseries\color{teaiblue}Abstract}
    \vskip 0.3em
  \end{center}
  Egocentric human video offers a scalable alternative to robot data for pretraining, yet models pretrained on such video consistently underperform those pretrained on robot data. We attribute this gap to a missing signal, the active perception behavior in egocentric videos, where humans continuously reposition their viewpoint during manipulation, inducing camera motion that standard pipelines treat as noise.
To address this, we present \textbf{ActiveMimic}, a pretraining framework that recovers synchronized camera and wrist trajectories from a single body-worn RGB camera, models camera motion as a viewpoint action, and jointly learns active perception and manipulation from in-the-wild egocentric human video before adapting to a target robot.
Empirically, real-world experiments across tasks with diverse active perception demands show that ActiveMimic consistently surpasses baselines pretrained on human video and matches state-of-the-art models pretrained on robot data.
Further analysis provides evidence that active perception capability originates from egocentric human video pretraining rather than robot-specific fine-tuning, confirming active perception as the key to unlocking egocentric human video for robot pretraining.
\par
  \vskip 0.3em
  {\small\textbf{Project Page:} \url{https://activemimic.github.io/}}
\end{tcolorbox}

\newpage
\section{Introduction}
Robot foundation models have become a central paradigm in robotic manipulation~\cite{rt2,openvla,pi0,gr00t,rdt,pi05}. A common training strategy combines a Vision-Language Model (VLM) with an action expert~\cite{dit,flow}, pretrains on large-scale robot data~\cite{oxe}, and adapts to downstream tasks. However, robot data remains expensive to collect, difficult to scale, and limited in task diversity. Instead, egocentric human videos offer a scalable alternative, being cheaper to acquire, covering a broader range of daily activities, and are easy to scale. While appealing, models pretrained on egocentric human data consistently underperform those pretrained on robot data.

Existing studies attribute this gap to the absence of action supervision and focus on constructing proxy action labels, such as hand trajectories~\cite{immimic,in-n-on}, hand point clouds~\cite{motovla}, or object motion signals~\cite{developing}. These approaches, however, miss a key signal: during manipulation, humans continuously reposition their viewpoint through head and body movements, inducing substantial camera motion in egocentric videos that standard pipelines treat as noise. In this paper, we argue that explicitly modeling this active perception behavior~\cite{active-perception,revisiting,active-vision} is key to unlocking egocentric human video for robot pretraining.

More specifically, modeling active perception requires recovering synchronized camera and wrist trajectories from egocentric human videos. However, wrist motion recovered from such videos inevitably conflates hand movement with camera rotation and translation, resulting in an inherent camera and hand coupling. Without resolving this coupling, a model cannot correctly learn either camera motion or hand motion. While existing methods that decouple camera motion and hand motion rely on dedicated capture hardware beyond a single body-worn RGB camera~\cite{egovla,emma,egohumanoid,humanoid}, preventing them from scaling to in-the-wild video, our goal is to resolve this coupling without specialized hardware, computing synchronized camera and wrist trajectories using off-the-shelf vision models alone and producing a unified action representation that captures how perception and manipulation jointly evolve.

With this in mind, we introduce \name{}, a pretraining framework that models viewpoint and wrist motion so as to perceive and act in an active manner. In particular, \name{} derives a unified action representation encoding the viewpoint motion of the camera alongside the bimanual wrist motion, all expressed in a common reference frame, allowing the model to learn their relationships through a single flow matching objective. We compute this unified action space on Ego4D~\cite{ego4d}, a large-scale egocentric dataset covering diverse daily activities and hand-object manipulation. It is worth noting that the approach is general and can be extended to any in-the-wild egocentric data. Once camera and wrist actions are aligned, we pretrain the model to predict both camera and wrist actions from egocentric observations, learning active perception jointly with manipulation. Finally, the pretrained model is adapted to the target robotic embodiment using robot-specific data, transferring the active perception capability acquired during pretraining.

Real-world experiments on tasks spanning diverse active perception demands show that \name{} consistently surpasses baselines pretrained on human video and matches state-of-the-art models pretrained on robot data. Our analysis further reveals that active perception originates from egocentric video pretraining rather than robot-specific fine-tuning, and that camera motion supervision facilitates representational transfer from human perception to robot control.

In summary, our contributions are threefold. \textbf{(a) \name{}: an active-perception-aware pretraining framework for in-the-wild egocentric video.} We extract synchronized camera and wrist trajectories from egocentric human video and jointly model active perception with manipulation, enabling scalable pretraining without dedicated capture hardware. \textbf{(b) Active perception is the key to unlocking egocentric human video for robot pretraining.} Real-world experiments demonstrate that camera motion supervision consistently improves success rate across tasks with diverse active perception demands. \textbf{(c) Active perception originates from pretraining and transfers from human to robot.} We show that active perception is acquired during egocentric pretraining rather than robot fine-tuning, and that camera motion supervision facilitates representational transfer from human perception to robot control.

\section{Related Work}

\paragraph{Learning from human videos}
Human videos offer a cheaper, more scalable, and more diverse alternative to robot data for pretraining. One line of work estimates proxy action labels from such videos, including hand trajectories~\cite{immimic,in-n-on}, hand point clouds~\cite{motovla}, and object motion signals~\cite{developing}, but supervises only hand or object motion and offers no signal about viewpoint action. A complementary line reads both viewpoint and hand actions directly from dedicated capture hardware~\cite{egovla,emma}, restoring action supervision at the cost of additional cameras~\cite{being} or wearable sensors~\cite{egohumanoid,humanoid,egomimic} beyond a single body-worn RGB camera, which restricts its applicability to in-the-wild egocentric human video.

\paragraph{Active perception}
A long-standing problem in robotics and computer vision is active perception~\cite{active-perception,revisiting,active-vision}, where an agent actively controls its viewpoint to reduce perceptual uncertainty rather than passively receiving images. Classically, it has been studied as Next-Best-View planning~\cite{receding,closed,determination,autonomous,online,affordance}, where viewpoint selection is optimized independently of downstream manipulation. Recent work instead models camera motion and manipulation actions jointly within a shared action space, learning perception and action end-to-end~\cite{egomi,via,activeumi,active-vision-need, eye, television,sapave}. This new paradigm, however, relies on dedicated data collection with human-operated capture rigs, ranging from VR headsets and controllers for robot teleoperation~\cite{activeumi,egomi} to wearable devices that record head and hand poses~\cite{egohumanoid}, and therefore cannot leverage in-the-wild egocentric human videos as a web-scale, organically produced corpus, analogous to the readily available web data that has driven the rapid scaling of modern VLM and LLM pretraining.

\paragraph{Learning active perception from egocentric human videos}
Among models trained on egocentric human videos~\cite{egovla,pi05-ego,scalable}, the dominant line supervises only proxy hand or object labels and leaves active perception unmodeled, while work that learns active perception from human data relies on additional cameras~\cite{being} or wearable sensors~\cite{egohumanoid,humanoid} beyond a single body-worn RGB camera rather than on in-the-wild egocentric video. In contrast, \name{} introduces a purely vision-based approach that recovers camera and wrist trajectories jointly from a single body-worn RGB camera, striking a balance between fidelity and scalability that enables active perception to be trained together with manipulation in a unified action space on in-the-wild egocentric videos.

\section{Method}
\begin{figure}[t]
    \centering
    \includegraphics[width=\linewidth]{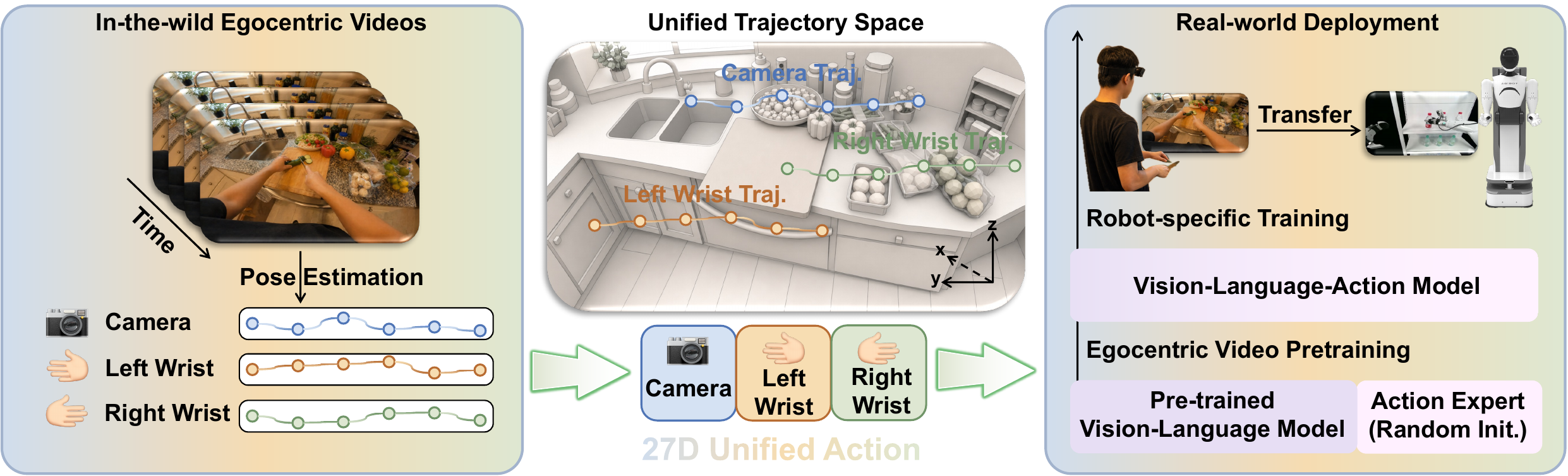}
    \caption{\textbf{Overview of \name{}.}
    \textit{Left:} recovering synchronized camera and wrist trajectories from a single body-worn RGB camera.
    \textit{Middle:} resolving camera-wrist coupling and encoding as a unified 27D action.
    \textit{Right:} pretraining on the 27D action to jointly model active perception and manipulation, then adapting to the target robot.}
    \vspace{-1.5em}
    \label{fig:framework}
\end{figure}

\name{} reframes egocentric human video pretraining around the coupled evolution of active perception and manipulation. Rather than treating egocentric camera motion as incidental noise, we interpret it as a viewpoint action that reflects how humans actively position their viewpoint during task execution. Starting from raw egocentric videos, we recover temporally aligned camera and wrist trajectories and represent them in a unified trajectory space, enabling joint modeling of active perception and manipulation (\Cref{subsec:dataset}). The model is then trained on this structured signal to predict both camera and wrist action from egocentric observations, enabling it to acquire transferable perceptual representations prior to adaptation to the target robotic embodiment (\Cref{subsec:model}).

\subsection{From Egocentric Video to Unified Action Space}
\label{subsec:dataset}
\vspace{-0.5em}
From a single body-worn RGB camera, we recover synchronized camera and wrist trajectories that jointly describe active perception and manipulation, without requiring additional sensors or controlled capture conditions. This involves three steps: recovering camera and wrist trajectories from RGB frames using off-the-shelf vision models, resolving the camera and wrist coupling by re-expressing all poses in a common reference frame, and encoding the result as a unified 27-dimensional action representation. In particular, we consider Ego4D~\cite{ego4d}, a large-scale egocentric dataset covering diverse daily activities; details of dataset filtering, temporal segmentation, and instruction annotation are provided in \Cref{subsec:appendix-filtering}.

\vspace{-0.7em}
\paragraph{Recovering camera and wrist trajectories.}
For each egocentric video, we estimate three synchronized pose trajectories using off-the-shelf vision models: the egocentric camera trajectory and the left and right wrist trajectories. We denote by $T^{\mathrm{tgt}}_{\mathrm{ref}} \in SE(3)$ the rigid transformation of the target frame expressed in the reference frame. For each frame $k \in \{1, \ldots, K\}$ in an episode of $K$ frames, we estimate the egocentric camera pose $T^{\mathrm{cam}_k}_{\mathrm{cam}_1}$, expressed in the coordinate system of the camera at the first frame of the episode, together with the left and right wrist poses $T^{\mathrm{wrist}^L_k}_{\mathrm{cam}_k}$ and $T^{\mathrm{wrist}^R_k}_{\mathrm{cam}_k}$, expressed in the coordinate system of the current-frame camera. The egocentric camera trajectory serves as the operational realization of the viewpoint action introduced earlier; rigidly attached to the wearer, it encodes active perception independent of the mounting configuration (head-, chest-, or glasses-mounted). Wrist poses are estimated by SAM-3D-Body~\cite{sam}. The camera trajectory is recovered by VGGT~\cite{vggt} as a scale-normalized path $\tilde{T}^{\mathrm{cam}_k}_{\mathrm{cam}_1}$, whose translational component is determined only up to a global scale factor. To recover the metric scale, we align the per-pixel depth maps from VGGT with metric depth estimates from UniDepth~\cite{unidepth} via a median depth ratio, aggregated into an episode-level scale factor $\lambda$. The metric camera trajectory $T^{\mathrm{cam}_k}_{\mathrm{cam}_1}$ is obtained by scaling the translational component of $\tilde{T}^{\mathrm{cam}_k}_{\mathrm{cam}_1}$ by $\lambda$ while keeping its rotation unchanged. Details of the scale recovery procedure are provided in \Cref{subsec:appendix-scale}.

\vspace{-0.7em}
\paragraph{Resolving camera and wrist coupling.}
The recovered camera and wrist trajectories are coupled: wrist poses are expressed in the current-frame camera coordinate system while the camera trajectory is anchored to the first frame, so any displacement in the wrist poses between frames reflects both actual wrist movement and the camera's own rotation and translation; using these wrist poses directly as action supervision would therefore conflate wrist movement with camera motion. We resolve this coupling by re-expressing all poses in a common chunk-relative reference frame. Since the policy operates on fixed-length temporal chunks rather than full episodes, we re-center all poses for each chunk. For a chunk of length $H$, let $i$ denote its start frame in the episode-level index and let $\tau \in \{0,\ldots,H-1\}$ denote the chunk-local offset. The corresponding episode-level frame index is $k=i+\tau$. We re-center all camera poses in this chunk to the coordinate system $\mathrm{cam}_i$ of the chunk's first frame. The chunk-relative camera pose at offset $\tau$ is
\begin{equation}
    T^{\mathrm{cam}_{i+\tau}}_{\mathrm{cam}_i} = \left(T^{\mathrm{cam}_i}_{\mathrm{cam}_1}\right)^{-1} T^{\mathrm{cam}_{i+\tau}}_{\mathrm{cam}_1},
\end{equation}
and the wrist poses are followed by composing the chunk-relative camera pose at offset $\tau$ with the current-frame wrist estimates,
\begin{equation}
    T^{\mathrm{wrist}^L_{i+\tau}}_{\mathrm{cam}_i} = T^{\mathrm{cam}_{i+\tau}}_{\mathrm{cam}_i}\, T^{\mathrm{wrist}^L_{i+\tau}}_{\mathrm{cam}_{i+\tau}},
\end{equation}
and analogously for the right wrist. This construction decouples camera and wrist motions by placing them in a single spatial reference frame $\mathrm{cam}_i$.

\vspace{-1.0em}
\paragraph{27D action representation.}
The decoupled chunk-relative poses are encoded into a unified 27-dimensional action vector that jointly captures viewpoint action and bimanual manipulation. Each chunk-relative pose, written in homogeneous form as
\begin{equation}
    T = \begin{bmatrix} R & t \\ \mathbf{0}^\top & 1 \end{bmatrix}, \qquad R \in SO(3), \; t \in \mathbb{R}^3,
\end{equation}
is encoded by its translation and a continuous 6D rotation representation~\cite{rotate6d}:
\begin{equation}
    p = t \in \mathbb{R}^3, \qquad r_{6D} = \bigl[R_{:,1};\, R_{:,2}\bigr] \in \mathbb{R}^6,
\end{equation}
where $R_{:,j}$ denotes the $j$-th column of $R$. Concatenating the camera and both wrist encodings yields a unified chunk-relative action vector for each chunk start $i$ and offset $\tau$:
\begin{equation}
    a_{i,\tau} = \bigl[\underbrace{p^{\mathrm{cam}}_{i,\tau},\,r^{\mathrm{cam}}_{6D,i,\tau}}_{\text{camera (9D)}},\,\underbrace{p^{\mathrm{wrist}^L}_{i,\tau},\,r^{\mathrm{wrist}^L}_{6D,i,\tau}}_{\text{left wrist (9D)}},\,\underbrace{p^{\mathrm{wrist}^R}_{i,\tau},\,r^{\mathrm{wrist}^R}_{6D,i,\tau}}_{\text{right wrist (9D)}}\bigr] \in \mathbb{R}^{27}.
\end{equation}
This unified 27D action space enables the model to jointly learn the coupled dynamics of camera and wrist motion within a single prediction objective.

\vspace{-0.8em}
\subsection{Architecture and Training Strategy}
\label{subsec:model}
\vspace{-0.5em}
With the decoupled camera and wrist action from \Cref{subsec:dataset}, we introduce a two-stage training strategy that injects active perception capability into the model. We first describe the model architecture, then detail the training strategy.

\vspace{-1.0em}
\paragraph{Architecture and training objective.}
The architecture of \name{} adopts a mix-of-transformers design~\cite{mot,bagel,pi0} that combines a visual-language prefix with an action-expert suffix. The visual-language prefix encodes images and a tokenized prompt into a multimodal context, onto which the action expert attends together with the current state and a continuous time variable to predict a chunk of future continuous actions. The policy is trained with a conditional flow-matching objective. The loss is defined as
\begin{equation}
    \mathcal{L} = \mathbb{E}_{a, \epsilon, t} \bigl\| v_t(a_t, o) - (\epsilon - a) \bigr\|_2^2,
\end{equation}
where $a_t = t\epsilon + (1-t)a$ is a noisy sample of the clean action chunk $a$, Gaussian noise $\epsilon \sim \mathcal{N}(0, I)$, time step $t \sim \mathcal{U}(0, 1)$, and $o$ denotes the overall conditioning context. The action chunk $a$ refers to the 27D unified action defined in \Cref{subsec:dataset} during egocentric human video pretraining and to the robot action chunk during robot-specific fine-tuning. At inference, the prefix representation is encoded once and cached, and the action chunk is recovered by initializing from Gaussian noise and iteratively denoising via Euler integration along the learned velocity field.

\vspace{-1.0em}
\paragraph{Two-stage training.}
Training follows a two-stage recipe: an egocentric human video pretraining stage on the dataset constructed in \Cref{subsec:dataset}, followed by a robot-specific training stage that adapts the pretrained policy to the target robotic embodiment. During pretraining, the visual-language prefix is initialized from a pretrained VLM checkpoint~\cite{paligemma} while the action expert is initialized at random, and the policy is supervised with the chunk-relative camera and wrist targets, so that it learns to model active perception jointly with manipulation from large-scale egocentric human video. The subsequent robot-specific training stage retains the same architecture and is initialized entirely from the pretrained weights, training on robot-specific data to transfer the active perception capability acquired during pretraining to the robotic embodiment.

\vspace{-1.0em}
\section{Experiments}
\label{sec:result}
\begin{figure*}[t]
    \centering
    \includegraphics[width=\linewidth]{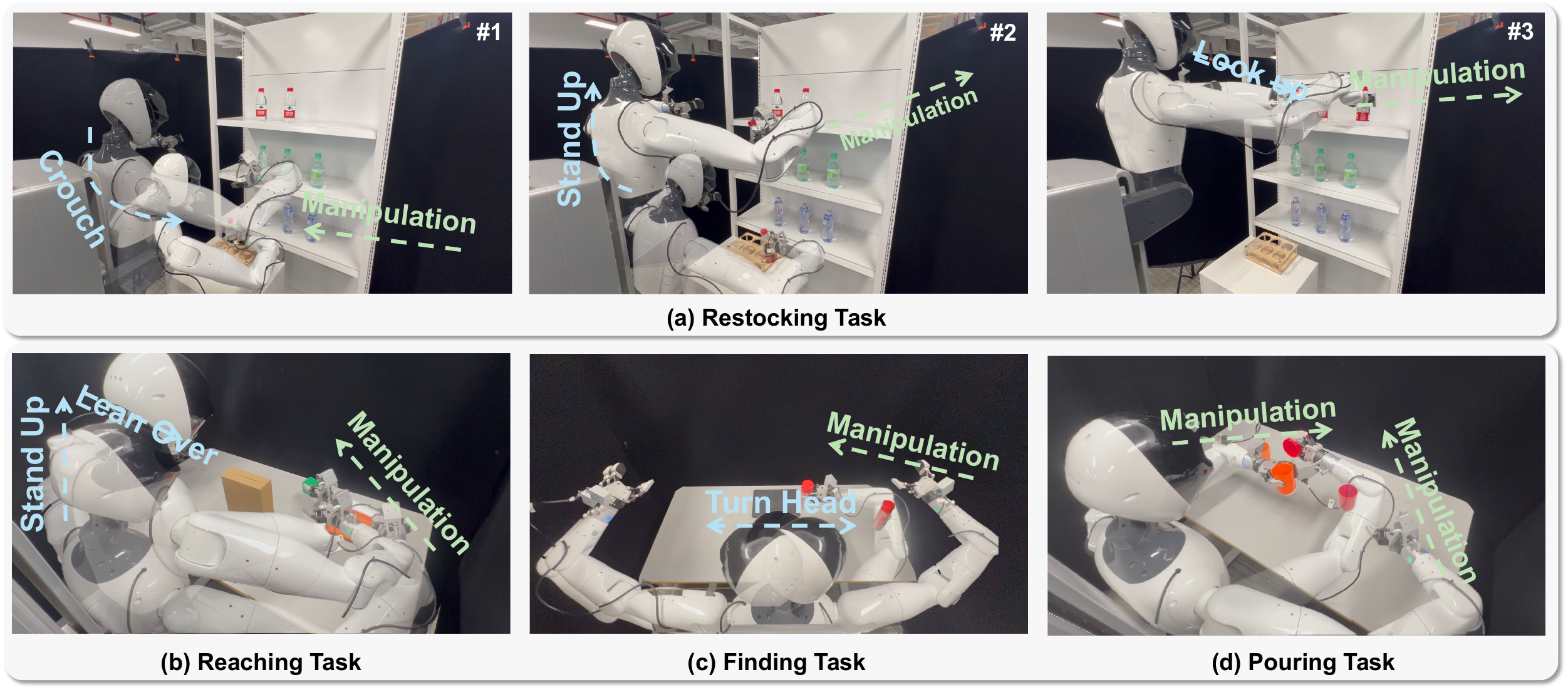}
    \caption{\textbf{Real-world tasks.}
    \textit{(a) Restocking}: the robot crouches to pick up a water bottle from the table, then stands
    and looks up to scan the shelf for an empty slot and places it.
    \textit{(b) Reaching}: the robot stands up and leans over an obstacle to reach the target
    object behind it.
    \textit{(c) Finding}: the robot turns its head left or right to locate a yogurt and
    grasps it with the corresponding arm.
    \textit{(d) Pouring}: the robot uses both hands to transfer liquid from a source container
    to a receiving container.}
    \label{fig:task}
\end{figure*}

We structure our evaluation around four questions that together assess whether
active perception is the key to unlocking egocentric human video for robot
pretraining.
\textbf{Q1} (\Cref{subsec:exp-main-results}). Does camera motion supervision improve real-world task performance?
\textbf{Q2} (\Cref{subsec:exp-dataset}). Do the camera and wrist trajectories recovered from egocentric video carry effective pretraining signals?
\textbf{Q3} (\Cref{subsec:exp-head-ablation}). Does active perception come from egocentric pretraining, and how does the model use it?
\textbf{Q4} (\Cref{subsec:exp-view-transfer}). Does camera motion supervision enable human-to-robot representational transfer?

\subsection{Experimental Setup}
\label{subsec:exp-setup}
\paragraph{Robot platform.}
We conduct all real-world experiments on a humanoid upper-body robot (AGIBOT G1) equipped with a 2-DoF head, a 2-DoF waist, and two 7-DoF arms with parallel-jaw grippers. The robot observes through three RGB cameras: one head-mounted and two wrist-mounted. The head camera, together with the head and waist joints, forms the active perception subsystem that enables viewpoint repositioning during task execution.

\paragraph{Tasks.}
We evaluate \name{} on four real-world tasks spanning the active perception spectrum (\Cref{fig:task}). \textit{(a) Restocking} is the most demanding: the robot crouches to pick up a water bottle from the table, stands and looks up to scan the shelf for an empty slot, then places the bottle. The shelf has three tiers at 70, 100, and 130\,cm; we award one point for pickup and one for placement. \textit{(b) Reaching} requires standing up and leaning over a 24\,cm obstacle to grasp a target object initialized in a 20$\times$20\,cm region behind it. \textit{(c) Finding} requires active search: the target yogurt is initialized in one of two 15$\times$25\,cm regions on the left or right side of the table, and the robot turns its head to locate and grasp it with the corresponding arm.
\textit{(d) Pouring} requires bimanual coordination to transfer liquid between two containers initialized in separate 20$\times$20\,cm regions. We train the four tasks on 270, 30, 60, and 90 teleoperated demonstrations and evaluate over 81, 18, 36, and 45 trials, respectively. We report end-to-end success rate as the primary metric and additionally score Restocking by average points per trial.

\paragraph{Pretraining data.}
We build our pretraining corpus from the Hands and Objects subset of Ego4D, which
already targets egocentric hand-object manipulation. We further filter this subset
to remove clips unsuitable for active perception supervision, yielding 2{,}561
episodes that amount to roughly 10 hours of video at 10\,fps and an average of ~130
frames per episode. Details of the additional filtering procedure are provided in
\Cref{subsec:appendix-filtering}.

\paragraph{Baselines.}
We compare \name{} against four baselines. (i) \emph{$\pi_0$}~\cite{pi0}, initialized from the publicly released checkpoint and fine-tuned on our robot-specific data. (ii) \emph{MotoVLA}~\cite{motovla}, a state-of-the-art model pretrained on human video whose pretraining corpus mixes robot data with RH20T~\cite{rh20t} human video, serving as the strongest available representative of pretraining on human video. (iii) \emph{\name{}\textsubscript{wrist-only}} shares the Ego4D~\cite{ego4d} corpus and architecture of \name{} but is supervised only with the 18D wrist action, isolating the contribution of camera motion supervision. (iv) \emph{\name{}\textsubscript{sft-only}} skips egocentric pretraining entirely and trains only on robot-specific data, isolating the contribution of egocentric human video pretraining.

\begin{figure*}[t]
    \centering
    \includegraphics[width=\linewidth]{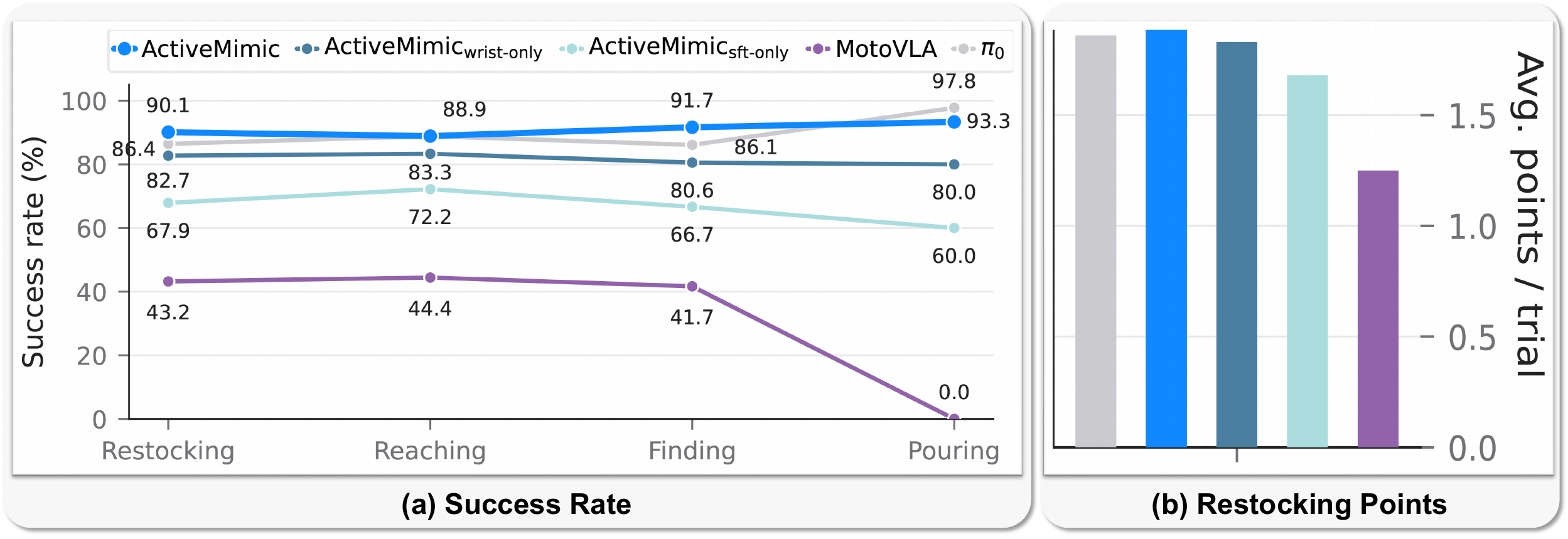}
    \caption{\textbf{Real-world results.}
    \textit{(a) Success rate}: end-to-end success rate (\%) on the four real-world tasks.
    \textit{(b) Restocking points}: average points per trial on Restocking, with one point awarded for picking up the bottle and one for placing it on the shelf.}
    \label{fig:main_results}
\end{figure*}

\subsection{Comparison with Baselines}
\label{subsec:exp-main-results}
\Cref{fig:main_results} shows that \name{} surpasses all baselines on all four
tasks, achieving success rates of 90.1\% on Restocking, 88.9\% on Reaching, 91.7\%
on Finding, and 93.3\% on Pouring.
Among the \name{} variants, both \name{}\textsubscript{wrist-only}
and \name{}\textsubscript{sft-only} fall behind \name{} across the board, confirming
that camera motion supervision during egocentric pretraining is the key differentiating factor. MotoVLA, which
leverages a large mixed corpus of robot and human data, also falls behind
\name{} by a substantial margin on all tasks.
Beyond these baselines, \name{} achieves comparable or higher
success rates than $\pi_0$ on all four tasks, showing that egocentric video
pretraining matches a state-of-the-art model pretrained on robot data. On Restocking
and Finding, the two tasks with the highest active perception demands, \name{} clearly
surpasses $\pi_0$ (90.1\% vs.\ 86.4\% and 91.7\% vs.\ 86.1\%),
indicating that egocentric video provides active perception advantages that robot data
alone does not capture.
We further investigate where this capability originates (\Cref{subsec:exp-head-ablation}) and whether it transfers from human to robot (\Cref{subsec:exp-view-transfer}).

\begin{figure}[t]
    \centering
    \includegraphics[width=0.99\linewidth]{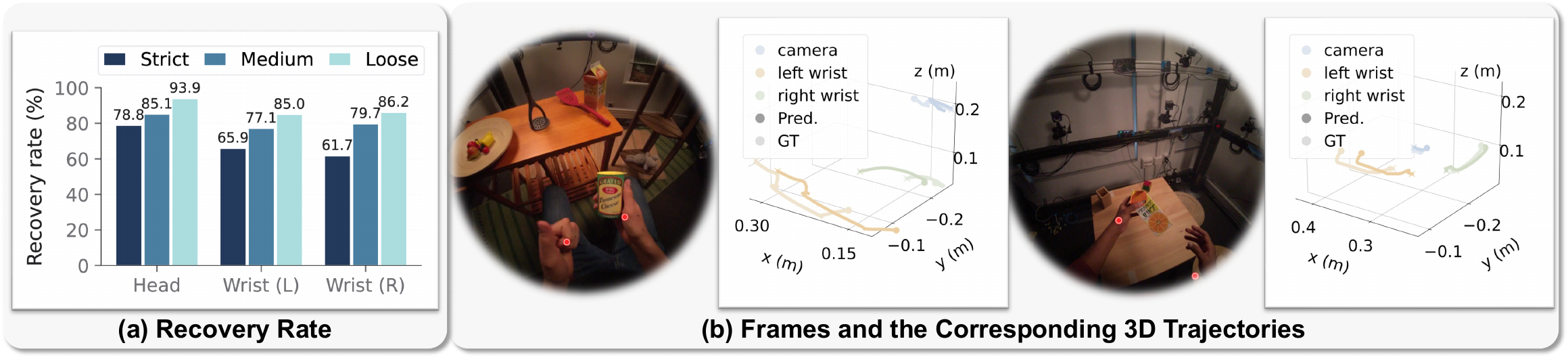}
    \caption{\textbf{Dataset characterization.}
    \textit{Left}: recovery rates of predicted head and wrist poses on HOT3D at three tolerance tiers.
    \textit{Right}: for two HOT3D videos, predicted wrist projections on a sampled frame and 3D chunk trajectories starting from that frame.}
    \label{fig:dataset}
\end{figure}

\subsection{Egocentric Video Yields Effective Pretraining Labels}
\label{subsec:exp-dataset}
The 27D action labels constructed in \Cref{subsec:dataset} are designed to expose the
coupling structure between active perception and manipulation from in-the-wild egocentric
video. Validating this design requires an egocentric dataset with ground-truth head and
wrist pose annotations, which Ego4D itself does not provide. We therefore evaluate our approach on HOT3D~\cite{hot3d}, an external
egocentric dataset that supplies such annotations.
We quantify label fidelity through the recovery rate on a randomly sampled 10\% subset of HOT3D videos: for each sampled frame, the estimated head and wrist poses are compared against ground-truth annotations, and a frame is considered recovered when both the translational error and the rot6d L2 error fall within a specified tolerance. Under the strict tier ($\text{pos} \leq 0.8\,\text{m}$, rot6d L2 $\leq 0.6$), head recovery reaches 78.82\%, with left and right wrist recovery at 65.93\% and 61.72\%, respectively; under the loose tier, all three body parts exceed 85\% (Fig.~\hyperref[fig:dataset]{\ref*{fig:dataset}a}). The approach operates purely from RGB video, without motion capture, inertial sensors, or calibrated multi-camera rigs, a deliberate fidelity-vs-scalability design choice that allows it to scale to arbitrary in-the-wild egocentric video. In addition, qualitative results show that estimated trajectories closely follow ground-truth trends on sampled HOT3D episodes (Fig.~\hyperref[fig:dataset]{\ref*{fig:dataset}b}). Together, these results (Fig.~\hyperref[fig:dataset]{\ref*{fig:dataset}}) confirm that the labels carry effective pretraining signals.

\begin{figure*}[t]
    \centering
    \includegraphics[width=\linewidth]{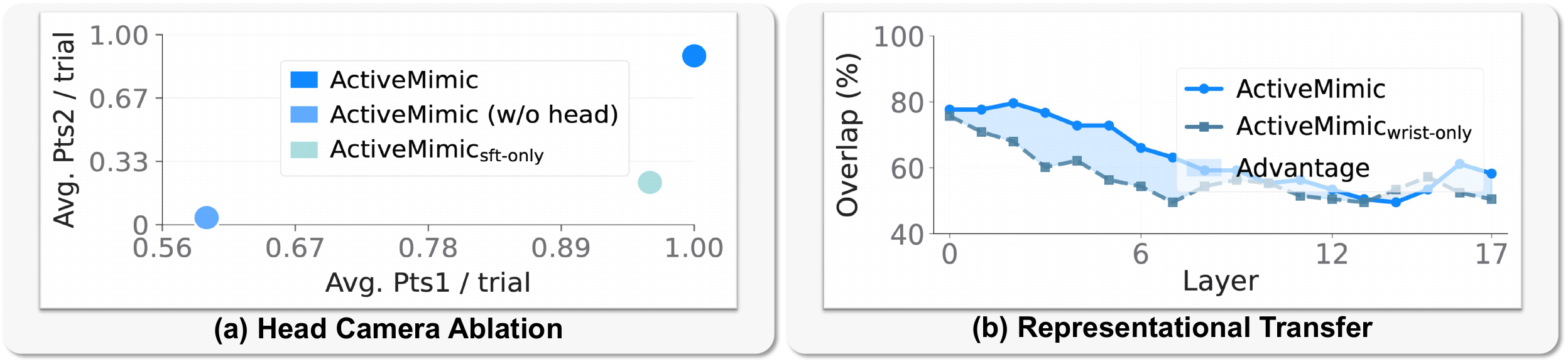}
    \caption{\textbf{Analysis experiments.}
    \textit{(a)} Scores on Restocking for crouching to grasp the bottle (Pts1) and looking up to place it (Pts2).
    \textit{(b)} Per-layer overlap (\%) of the top-10\% activated units under head-view vs.\ full-view inputs for \name{} and \name{}\textsubscript{wrist-only}.}
    \label{fig:ablation}
\end{figure*}

\subsection{The Head Camera Enables Pretrained Active Perception}
\label{subsec:exp-head-ablation}
To pinpoint whether active perception comes from egocentric pretraining and how the model deploys it, we ablate Restocking under three inference conditions (Fig.~\hyperref[fig:ablation]{\ref*{fig:ablation}a}): \name{} with all three cameras, \name{} with the head camera zeroed out (w/o head), and \name{}\textsubscript{sft-only} with all cameras. Notably, all three conditions reliably complete the pickup point, but the placement point reveals a stark separation. \name{} scores 24 out of 27 on placement, whereas \name{}\textsubscript{sft-only} achieves only 6 out of 27. This fourfold gap indicates that active perception capability is acquired during egocentric pretraining rather than robot-specific fine-tuning. Removing the head camera from the pretrained model collapses placement further to 1 out of 27, confirming that the model realizes this capability through the head camera. Together, egocentric pretraining provides active perception capability, and the head camera is how the model uses it.

\subsection{Human-to-Robot Representational Transfer via Camera Motion Supervision}
\label{subsec:exp-view-transfer}
To investigate how camera motion supervision facilitates human-to-robot transfer, we compare \name{} and \name{}\textsubscript{wrist-only} under two inference conditions: full-view, where the model receives all three cameras, and head-view, where it receives only the head camera, approximating the single egocentric viewpoint in pretraining. Specifically, for each layer in the action expert, we identify the top-$K$\% most activated units under each view and compute their overlap. The resulting overlap between head-view and full-view thus measures how much of the egocentric representational structure is preserved under the robot's multi-camera observation. Because this egocentric structure is learned from human video pretraining, higher preservation directly reflects stronger human-to-robot transfer. As shown in Fig.~\hyperref[fig:ablation]{\ref*{fig:ablation}b}, \name{} maintains consistently higher overlap than \name{}\textsubscript{wrist-only} across the early-to-mid layers (layers 0 through 11), where perceptual representations are encoded~\cite{look}. The higher overlap in perceptual layers indicates that camera motion supervision produces representations more robust to the observation modality shift, providing representational-level evidence that camera motion supervision strengthens human-to-robot transfer. We report $K{=}10$ and show in \Cref{subsec:appendix-k-sensitivity} that the conclusion is robust to the choice of $K$.

\vspace{-1.0em}
\section{Conclusion}
We introduce \name{}, an active-perception-aware pretraining framework for in-the-wild egocentric video. Across real-world tasks, \name{} consistently surpasses baselines pretrained on human video, confirming active perception as the key to unlocking egocentric human video for robot pretraining. We further provide evidence that active perception originates from egocentric pretraining and that camera motion supervision facilitates representational transfer from human perception to robot control.
Limitations are discussed in \Cref{sec:appendix-limitations}.

\clearpage

\bibliography{main}  %

\newpage
\appendix

\section{From Egocentric Video to Unified Action Space}
\label{sec:appendix-data}

\subsection{Metric Scale Recovery}
\label{subsec:appendix-scale}

The camera trajectory recovered by VGGT is a scale-normalized path $\tilde{T}^{\mathrm{cam}_k}_{\mathrm{cam}_1}$ whose translational component is determined only up to a global scale factor. To recover the metric scale, we align the per-pixel depth map $D^{\mathrm{norm}}_k$ from VGGT with the per-pixel metric depth map $D^{\mathrm{metric}}_k$ from UniDepth. A per-frame scale is first computed as the median depth ratio over valid pixels,
\begin{equation}
    \lambda_k = \operatorname{median}_{(u,v)\in\Omega_k} \frac{D^{\mathrm{metric}}_k(u,v)}{D^{\mathrm{norm}}_k(u,v)},
\end{equation}
where $\Omega_k$ denotes the set of pixels with valid positive depth values in both $D^{\mathrm{norm}}_k$ and $D^{\mathrm{metric}}_k$. The per-frame scales are then aggregated into an episode-level scale $\lambda = \operatorname{median}_{k \in \{1,\ldots,K\}} \lambda_k$.
The metric camera trajectory is then obtained by scaling only the translational component of the scale-normalized transform,
\begin{equation}
    \tilde{T}^{\mathrm{cam}_k}_{\mathrm{cam}_1}
    =
    \begin{bmatrix}
        R^{\mathrm{cam}_k}_{\mathrm{cam}_1} & \tilde{t}^{\mathrm{cam}_k}_{\mathrm{cam}_1} \\
        \mathbf{0}^\top & 1
    \end{bmatrix},
    \qquad
    T^{\mathrm{cam}_k}_{\mathrm{cam}_1}
    =
    \begin{bmatrix}
        R^{\mathrm{cam}_k}_{\mathrm{cam}_1} & \lambda\,\tilde{t}^{\mathrm{cam}_k}_{\mathrm{cam}_1} \\
        \mathbf{0}^\top & 1
    \end{bmatrix}.
\end{equation}

\subsection{Video Filtering and Segmentation}
\label{subsec:appendix-filtering}

As described in \Cref{subsec:dataset}, we identify hand-object manipulation segments from Ego4D through a two-stage filtering procedure that combines VLM-based temporal segmentation with LLM-based semantic filtering.

\paragraph{VLM-based temporal segmentation.}
For each Ego4D clip that passes an initial duration filter, a VLM (Qwen3-VL-8B-Instruct~\cite{qwen3-vl}) parses the full egocentric video and proposes candidate manipulation segments. The model is prompted to retain only intervals in which the camera wearer purposefully uses their hands to manipulate physical objects, excluding passive observation, walking, waiting, and pure camera motion. For each retained segment, the model outputs a start and end time, an action verb, a list of manipulated objects, and a natural-language task instruction composed from the action and objects. This task instruction serves as the language prompt during pretraining. Adjacent or overlapping segments with the same task description are merged, and a duration filter is applied to remove segments that are too short to contain meaningful manipulation or too long for efficient downstream processing. The full prompt is provided in \Cref{fig:prompt-vlm}.

\paragraph{LLM-based semantic filtering.}
These candidates are then filtered by an LLM (Qwen3-30B-A3B-Instruct~\cite{qwen3}) against three semantic criteria: the action must involve hand-object manipulation, the manipulated objects must be artificial physical objects, and the scene must be indoors. Segments involving body parts or other humans as targets, natural or outdoor materials, outdoor activities, or non-manipulation actions are removed. This stage produces the final set of high-confidence indoor manipulation segments that are sampled and extracted at 10\,fps for pose estimation. The full prompt is provided in \Cref{fig:prompt-llm}.

\begin{figure*}[t]
    \centering
    \includegraphics[width=\linewidth]{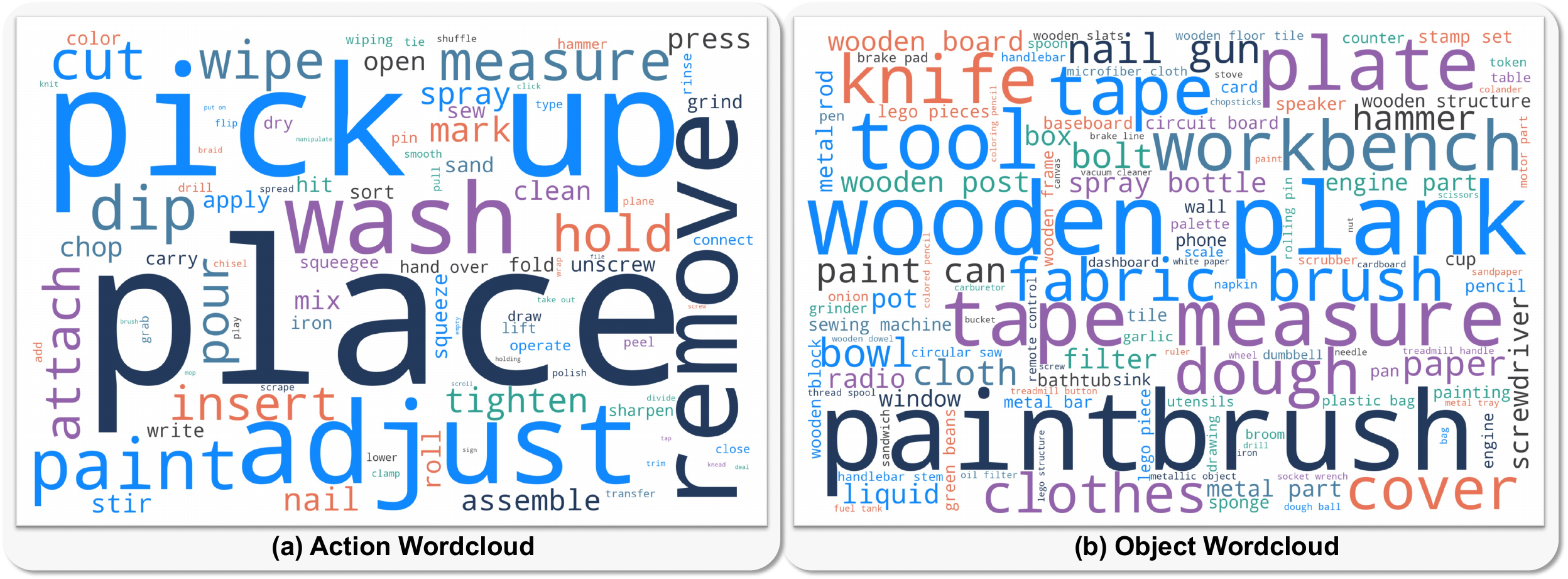}
    \caption{\textbf{Pretraining corpus statistics.} Word cloud of \textbf{(a)} action verbs and \textbf{(b)} manipulated objects in the final pretraining corpus after filtering, showing broad coverage of manipulation actions and object categories.}
    \label{fig:wordcloud}
\end{figure*}

\paragraph{Dataset statistics.}
\Cref{fig:wordcloud} visualizes the action verb and object noun distributions of the final pretraining corpus. The verb word cloud reflects the diversity of manipulation actions, while the noun word cloud shows the breadth of manipulated object categories, confirming that the filtering procedure preserves semantic variety suitable for general-purpose pretraining.

\section{Training Details}
\label{sec:appendix-training}

The model comprises a 3B visual-language prefix and a 0.6B action expert. As described in \Cref{subsec:model}, training follows a two-stage recipe. The egocentric human video pretraining stage is further divided into a warm-up phase and a full training phase. During warm-up, the visual-language prefix is frozen and only the action expert is trained, allowing the randomly initialized action expert to reach a reasonable operating point before joint optimization. The full training phase then unfreezes all parameters and trains the entire model end-to-end. The robot-specific training stage initializes from the pretrained checkpoint and fine-tunes on task-specific robot data. \Cref{tab:hyperparams} summarizes the hyperparameters for each phase. Both \name{} and \name{}\textsubscript{wrist-only} share the same training configuration; the only difference is that \name{}\textsubscript{wrist-only} is supervised with the 18D wrist action instead of the full 27D action.

\begin{table}[h]
\centering
\small
\begin{tabular}{lccc}
\toprule
 & \textbf{Stage 1: Warm-up} & \textbf{Stage 1: Train} & \textbf{Stage 2} \\
\midrule
Frozen modules & VLM & None & None \\
Batch size & 64 & 64 & 64 \\
Optimizer & AdamW & AdamW & AdamW \\
LR schedule & Cosine decay & Cosine decay & Cosine decay \\
Peak LR & $1 \times 10^{-4}$ & $1 \times 10^{-4}$ & $1 \times 10^{-4}$ \\
Final LR & $1 \times 10^{-5}$ & $1 \times 10^{-5}$ & $1 \times 10^{-5}$ \\
Warmup steps & 200 & 1{,}000 & 100 \\
Total steps & 2{,}000 & 500{,}000 & ${\sim}$5 epochs \\
Action horizon & 50 & 50 & 50 \\
\bottomrule
\end{tabular}
\caption{\textbf{Training hyperparameters for each phase.} Stage 2 trains for approximately 5 epochs on each task.}
\label{tab:hyperparams}
\end{table}

\section{Experimental Details}
\label{sec:appendix-exp}

\subsection{Task Setup}
\label{subsec:appendix-task-setup}

\Cref{tab:task-setup} consolidates the detailed specifications for each evaluation task. All four tasks are executed on the same robot platform described in \Cref{subsec:exp-setup}.

\definecolor{secheader}{rgb}{0.88,0.92,0.97}
\begin{table}[h]
\centering
\small
\begin{tabular}{lcccc}
\toprule
 & \textbf{Restocking} & \textbf{Reaching} & \textbf{Finding} & \textbf{Pouring} \\
\midrule
\rowcolor{secheader} \multicolumn{5}{l}{\textit{Object specifications}} \\
Target object & Water bottle & Cubic block & Yogurt & Cup \\
Object height (cm) & 17 & 5 & 16 & 9 \\
Object diameter/side (cm) & 5.5 & 5 & 5 & 7 \\
\midrule
\rowcolor{secheader} \multicolumn{5}{l}{\textit{Environment specifications}} \\
Shelf tier heights (cm) & 70 / 100 / 130 & -- & -- & -- \\
Shelf tier size (cm) & 92.5 $\times$ 35 & -- & -- & -- \\
Obstacle height (cm) & -- & 24 & -- & -- \\
Init. region (cm) & -- & 20 $\times$ 20 & 15 $\times$ 25 ($\times$2) & 20 $\times$ 20 ($\times$2) \\
\midrule
\rowcolor{secheader} \multicolumn{5}{l}{\textit{Training and evaluation}} \\
Demonstrations & 270 & 30 & 60 & 90 \\
Evaluation trials & 81 & 18 & 36 & 45 \\
Metric & SR + Pts & SR & SR & SR \\
\bottomrule
\end{tabular}
\caption{\textbf{Detailed task specifications.} Init.\ region denotes the randomized object initialization area; ($\times$2) indicates two separate regions. SR = success rate; Pts = average points per trial (1 for pickup + 1 for placement).}
\label{tab:task-setup}
\end{table}

\subsection{Robustness Evaluation}
\label{subsec:appendix-robustness}

\begin{figure*}[t]
    \centering
    \includegraphics[width=\linewidth]{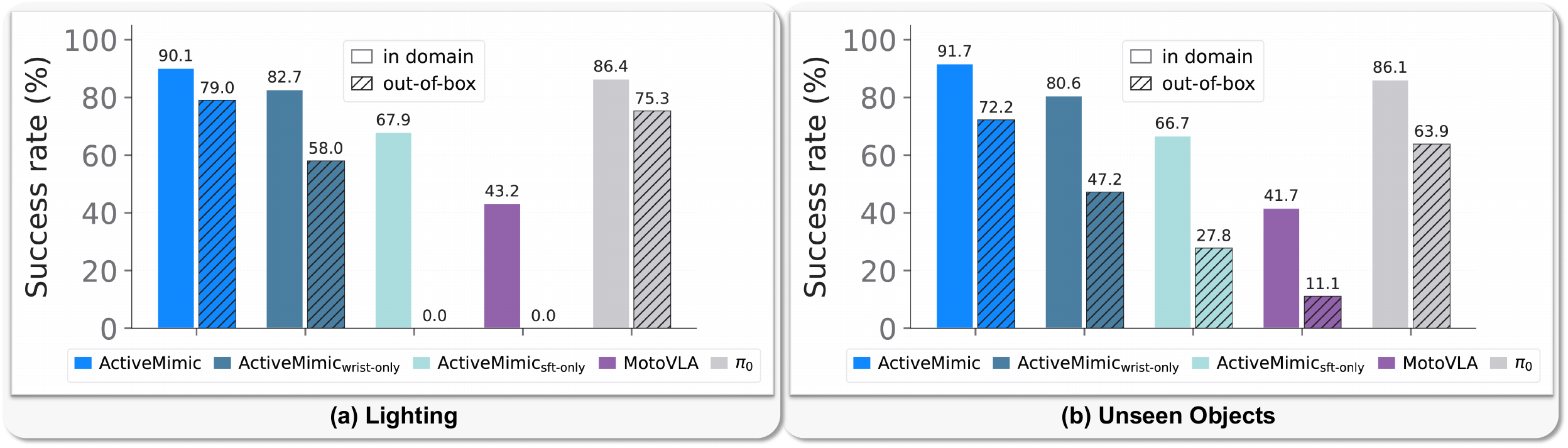}
    \caption{\textbf{Robustness evaluation.} \textbf{(a)} Restocking under alternating red/green/blue flashing light. \textbf{(b)} Finding with unseen yogurt objects (different packaging, identical shape and size). Solid bars denote in-domain (normal) conditions; hatched bars denote out-of-domain conditions. \name{} achieves the highest success rate under both perturbations and exhibits the smallest absolute drop among all models.}
    \label{fig:robust}
\end{figure*}

\paragraph{Restocking under flashing lighting.}
We evaluate all models on the Restocking task under alternating red/green/blue flashing light, using the same checkpoint as the main results, with 81 trials per condition. As shown in \Cref{fig:robust}(a), \name{} achieves the highest success rate under the flashing condition (79.0\%) and shows the smallest absolute drop among all models ($-11.1\%$ from 90.1\%). \name{}\textsubscript{wrist-only} drops by 24.7\% to 58.0\%, suggesting that camera motion supervision contributes to the robustness. \name{}\textsubscript{sft-only} and MotoVLA both collapse to 0\%, indicating that pretraining without active-perception-aware supervision provides no protection against visual perturbations. $\pi_0$ shows comparable robustness (75.3\%), dropping 11.1\%.

\paragraph{Finding with unseen objects.}
We replace the training yogurt with two unseen yogurt variants (different packaging, identical shape and size) and evaluate on the Finding task with 36 trials per condition. As shown in \Cref{fig:robust}(b), \name{} maintains the highest success rate among all models (72.2\%) and the smallest drop ($-19.5\%$ from 91.7\%). $\pi_0$ drops 22.2\% to 63.9\%. \name{}\textsubscript{wrist-only} drops 33.4\% to 47.2\%, while \name{}\textsubscript{sft-only} and MotoVLA drop to 27.8\% and 11.1\%, respectively. The larger gaps relative to the lighting experiment reflect that visual appearance is particularly load-bearing for object localization, yet \name{}'s active-perception pretraining provides the strongest generalization.

\begin{figure*}[t]
    \centering
    \includegraphics[width=\linewidth]{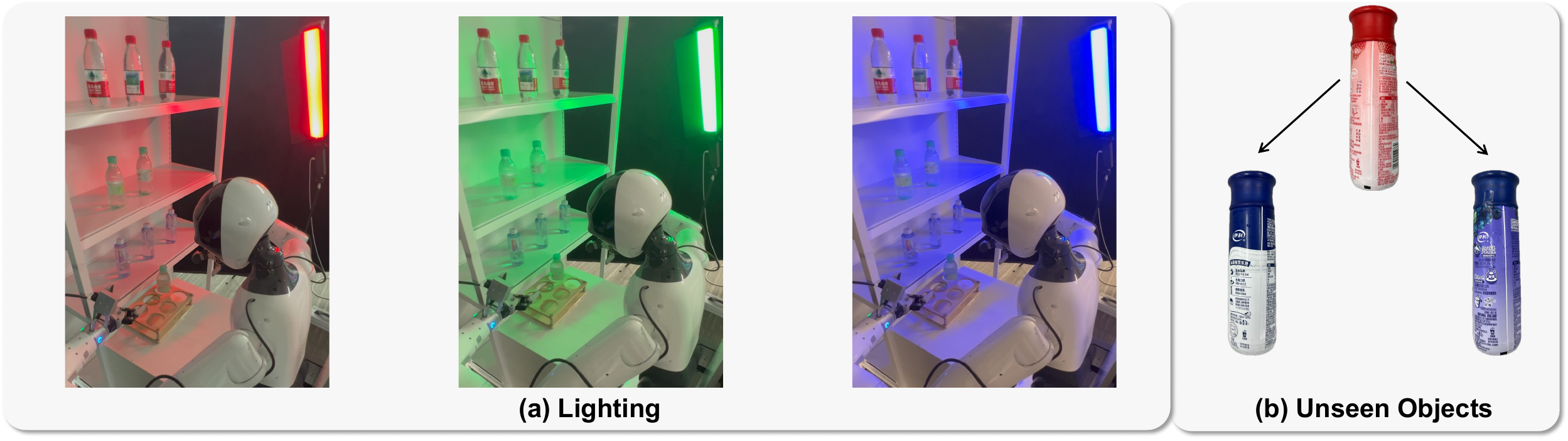}
    \caption{\textbf{Robustness evaluation setup.} \textbf{(a)} Restocking under alternating red, green, and blue flashing light. \textbf{(b)} Finding with two unseen yogurt variants (different packaging, identical shape and size) not present in training demonstrations. The training yogurt is shown at the top; the two unseen variants are shown below.}
    \label{fig:robust-setup}
\end{figure*}

\subsection{Failure Case Analysis}
\label{subsec:appendix-failure}

\begin{figure*}[t]
    \centering
    \includegraphics[width=\linewidth]{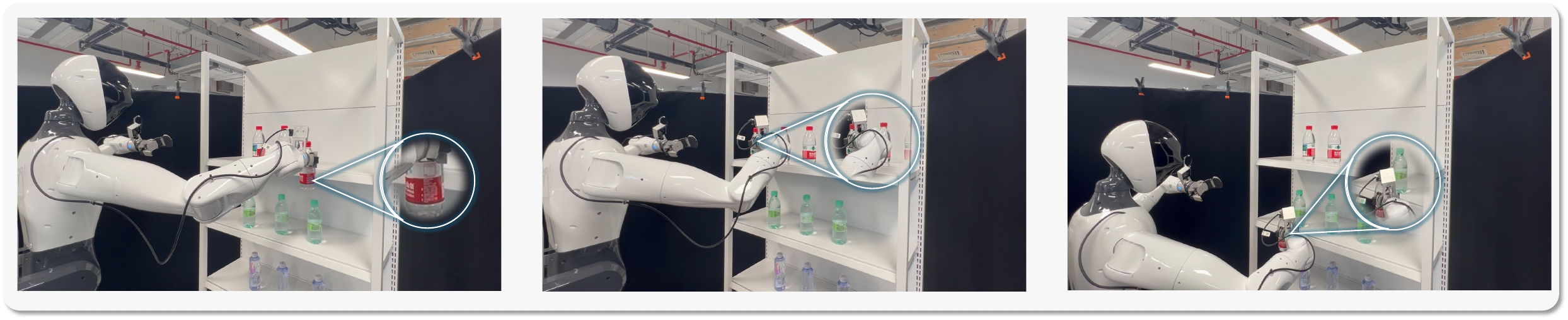}
    \caption{\textbf{Representative failure cases of \name{} without the head camera on Restocking.} All three failures occur at the placement point. From left to right: (1) correct shelf tier and lateral position, but the placement motion is imprecise and knocks over the shelf; (2) correct tier, wrong lateral position; (3) wrong tier entirely. All three stem from severing the visual loop that the pretrained model relies on to coordinate head and hand movements during active perception.}
    \label{fig:failure-case}
\end{figure*}

\Cref{fig:failure-case} presents representative failure cases of the \textit{w/o head} condition on Restocking. All three failures occur at the placement point. In the first case, the arm reaches the correct shelf tier and lateral position but the placement motion is imprecise and knocks over the shelf. In the second case, the arm places the bottle on the correct tier but at the wrong lateral position. In the third case, the arm targets the wrong tier entirely. All three failures stem from removing the head camera, which severs the visual loop that the pretrained model relies on to coordinate head and hand movements during active perception. Without this feedback, the head-hand coordination acquired during egocentric human video pretraining breaks down, producing increasingly coarse placement errors.

\subsection{Representational Transfer: K Sensitivity Analysis}
\label{subsec:appendix-k-sensitivity}

\begin{figure*}[t]
    \centering
    \includegraphics[width=\linewidth]{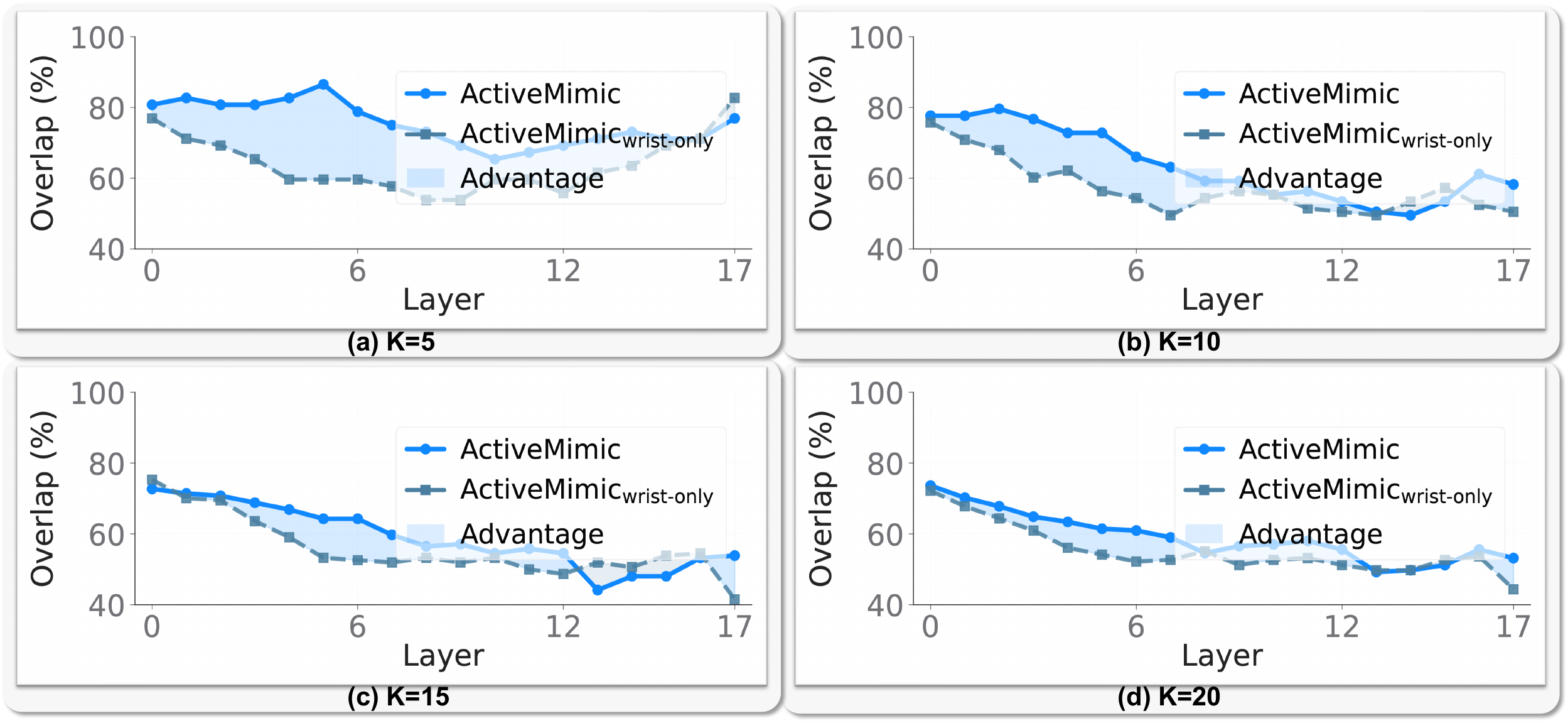}
    \caption{\textbf{K sensitivity analysis for representational transfer.} Top-$K$\% activation overlap between full-view and head-view inference conditions for \name{} and \name{}\textsubscript{wrist-only} across all action-expert layers, evaluated at $K=5, 10, 15, 20$. The shaded area indicates the advantage of \name{} over \name{}\textsubscript{wrist-only}. \name{} maintains consistently higher overlap across all $K$ values, confirming that the conclusion in \Cref{subsec:exp-view-transfer} is robust to the choice of $K$.}
    \label{fig:k-sensitivity}
\end{figure*}

\Cref{subsec:exp-view-transfer} reports representational transfer results at $K=10$. \Cref{fig:k-sensitivity} extends this analysis to $K \in \{5, 10, 15, 20\}$. Across all values of $K$, \name{} maintains consistently higher top-$K$\% activation overlap than \name{}\textsubscript{wrist-only}, confirming that the conclusion is robust to the choice of $K$.

\section{Limitations and Future Directions}
\label{sec:appendix-limitations}

\paragraph{Data scale.}
The current pretraining corpus comprises approximately 10 hours of filtered egocentric manipulation video from Ego4D. While this scale already yields significant gains over training from scratch, substantially larger egocentric corpora (~\eg, the full Ego4D~\cite{ego4d} or Ego-Exo4D~\cite{ego-exo}) are readily available and can be incorporated with the same automated procedure, which we expect to further strengthen the pretrained representations.

\paragraph{Embodiment diversity.}
All real-world experiments use a single humanoid platform. Because the pretraining stage is embodiment-agnostic, operating on egocentric video without any robot-specific input, extending to other embodiments only requires the robot-specific training stage with corresponding demonstrations. Validating this across a broader range of platforms is a natural next step.

\paragraph{Label fidelity.}
Action labels are obtained from vision-based pose estimation rather than hardware-recorded trajectories, which inevitably introduces estimation noise. Nonetheless, the pretrained models achieve strong downstream performance, suggesting that the learning objective is robust to moderate label noise. Label quality can be further improved as better off-the-shelf pose estimation methods become available.

\paragraph{Loco-manipulation.}
The current evaluation focuses on stationary tabletop and shelf manipulation. Extending \name{} to loco-manipulation on humanoid robots~\cite{egohumanoid,visualmimic,psi_0,wholebodyvla}, where the robot must coordinate locomotion and manipulation simultaneously, is a promising direction, as egocentric video datasets already contain abundant walking-while-manipulating footage that could serve as pretraining data.

\paragraph{Human-robot interaction.}
Egocentric video covers diverse daily activities, many of which naturally involve interactions with other people~\cite{holo,perceiving,ask,predicting}. Extending \name{} to human-robot interaction scenarios, where the robot must perceive and respond to human actions in shared workspaces, is a compelling direction that could leverage this inherent property of egocentric data.

\begin{figure*}[t]
\centering
\begin{minipage}{\textwidth}
\small
\begin{verbatim}
You are an expert in understanding egocentric videos involving hand-object interactions.

Please watch the entire egocentric video carefully and identify all time segments where
the camera wearer is performing a specific, goal-directed task that involves direct
interaction between their hands and a specific object.

A valid task must satisfy the following conditions:
- The person's hands are actively manipulating or interacting with a physical object
- The action has a clear purpose, such as "washing a dish", "opening a bottle", or
  "tightening a screw"
- Segments where the person is not using their hands to manipulate any object — such as
  walking, turning their head, looking around, standing still, observing, or waiting —
  should be excluded

The total duration of the video is [VIDEO_DURATION] seconds.

For each detected task segment, provide:
1. The start time (in seconds, integer only)
2. The end time (in seconds, integer only)
3. A concise description of the specific task being performed

Each description must include:
- The main manipulation action (a verb like "pick up", "place", "insert", "open", etc.)
- A list of one or more objects that are being manipulated
- A short natural language instruction generated from the action and objects

The segments may overlap in time if multiple tasks are performed in close succession
or simultaneously.

Return the results strictly in the following JSON format:
[
  {"start": 4, "end": 9, "action": "open", "objects": ["bag"], "task": "Open the bag"},
  {"start": 9, "end": 15, "action": "place", "objects": ["apple", "plate"],
   "task": "Place the apple on the plate"}
]
\end{verbatim}
\end{minipage}
\caption{\textbf{Prompt used for VLM-based temporal segmentation.} The model identifies manipulation segments from egocentric video and outputs structured annotations including a natural-language task instruction that serves as the language prompt during pretraining.}
\label{fig:prompt-vlm}
\end{figure*}

\begin{figure*}[t]
\centering
\begin{minipage}{\textwidth}
\small
\begin{verbatim}
You are a task filter for egocentric video clips.

You will be given a JSON object that represents a single video clip. Each clip contains
a list of task segments. Your job is to extract only segments that are suitable for
training a Vision-Language-Action (VLA) model focused on indoor hand-object manipulation.

Each segment includes:
- start, end: time in seconds
- action: verb describing the action
- objects: list of physical objects being interacted with
- task: natural language description

Filtering criteria — keep only segments that satisfy all three:
1. The action involves hand-object manipulation (e.g., pick up, cut, fold, assemble,
   insert, tighten, wipe, pour, etc.)
2. The object(s) must be artificial, physical items (tools, containers, utensils,
   electronics, furniture, fabric, household goods). Exclude: body parts (leg, hand,
   arm), people (man, woman, person), natural materials (plant, soil, mud, grass, tree).
3. The scene is likely indoors. Exclude: gardening, farming, outdoor repair, digging,
   planting, handling mud/branches/natural terrain.

Return a JSON object:
{"clip_uid": "...", "status": "success", "filtered_segments": [...]}
\end{verbatim}
\end{minipage}
\caption{\textbf{Prompt used for LLM-based semantic filtering.} The model retains only segments involving indoor hand-object manipulation of artificial objects.}
\label{fig:prompt-llm}
\end{figure*}

\end{document}